\documentclass[10pt,journal,cspaper,compsoc]{IEEEtran}
\ifCLASSOPTIONcompsoc
\else
\fi

\ifCLASSINFOpdf
   \usepackage[pdftex]{graphicx}
\else
\fi

\usepackage{wrapfig}

\usepackage{cite}
\usepackage{url}

\usepackage{amsmath}

\usepackage{color} 
\usepackage{multirow,times}

\usepackage{epstopdf} 

\mathchardef\mhyphen="2D

\usepackage{maplestd2e}

\usepackage{hyperref} 

\begin{document}

\bstctlcite{IEEEexample:BSTcontrol}

%
\title{Isomorphism between Differential and Moment Invariants under Affine Transform}

\author{Erbo Li and Hua Li
\IEEEcompsocitemizethanks{
\IEEEcompsocthanksitem E. Li is with EON Reality Inc, Irvine, CA 92618. Email: sophialiuli@gmail.com
\IEEEcompsocthanksitem H. Li is with Key lab of Intelligent Information Processing, Institute of Computing Technology, Chinese Academy of Sciences, 100190 Beijing; University of Chinese Academy of Sciences, 100049 Beijing. 
E-mail: lihua@ict.ac.cn
}
}


\markboth{}%
{Shell \MakeLowercase{\textit{Erbo Li et al.}}: Isomorphism between Differential and Moment Invariants under Affine Transform}

\IEEEcompsoctitleabstractindextext{%
\begin{abstract}

The invariant is one of central topics in science, technology and engineering. The differential invariant is essential in understanding or describing some important phenomena or procedures in mathematics, physics, chemistry, biology or computer science etc. The derivation of differential invariants is usually difficult or complicated.
This paper reports a discovery that under the affine transform, differential invariants have similar structures with moment invariants up to a scalar function of transform parameters. If moment invariants are known, relative differential invariants can be obtained by the substitution of moments by derivatives with the same order.
Whereas moment invariants can be calculated by multiple integrals, this method provides a simple way to derive differential invariants without the need to resolve any equation system.
Since the definition of moments on different manifolds or in different dimension of spaces is well established, differential invariants on or in them will also be well defined. Considering that moments have a strong background in mathematics and physics, this technique offers a new view angle to the inner structure of invariants. Projective differential invariants can also be found in this way with a screening process.


\end{abstract}

}

\maketitle

\IEEEdisplaynotcompsoctitleabstractindextext

%
\IEEEpeerreviewmaketitle

\section{Introduction}
\label{sec:intro}

The invariant is one of central topics in science, technology and engineering. The differential invariant is essential in understanding or describing some important phenomena or procedures in mathematics, physics, chemistry, biology or computer science etc. The derivation of differential invariants is usually difficult or complicated.

This paper reports a discovery that under the affine transform, differential invariants have similar structures with moment invariants up to a scalar function of transform parameters. If moment invariants are known, relative differential invariants can be obtained by the substitution of moments by derivatives with the same order.

Whereas moment invariants can be calculated by multiple integrals, this method provides a simple way to derive differential invariants without the need to resolve any equation system.

Since the definition of moments on different manifolds or in different dimension of spaces is well established, differential invariants on or in them will also be well defined. Considering that moments have a strong background in mathematics and physics, this technique offers a new view angle to the inner structure of invariants.

This paper is organized as follows: introduce the moment in section 2 and moment invariant in section 3; prove the moments relations under transforms in section 4; derive differential invariants from moment invariants in section 5; give extension to projective group in section 6 and finally in section 7 sum up some conclusions.

\section{Geometric moment and general linear transform}
\label{sec:2}

The idea of moment is simple and it has long history. In 2D we have the following.

\textbf{Definition 1: moment}

An order $(m+n)$ moment of an image $S$ is

\begin{equation}
\label{eq:1}
\mapleinline{inert}{2d}{M[mn] = int(int(x^m*y^n*rho(x, y), x), y)}{\[\displaystyle M_{{{\it mn}}}=\int \!\!\!\int \!{x}^{m}{y}^{n}\rho \left( x,y \right) {dx}\,{dy}\]}
\end{equation}

where: $m, n$ are positive integers; $ \rho \left( x,y \right) $ is a density function; the integral is defined on the image S.

\textbf{Definition 2: central movement}

The central movement is

\begin{equation}
\label{eq:2}
\mu _{{\it mn} } = \int \!\!\! \int \left(x - x' \right)^{m } \left(y - y' \right)^{n }\rho \left(x ,y \right) dx dy
\end{equation}

where $x'$ and $y'$ are the coordinates of the mass center or centroid,

\begin{equation}
\label{eq:3}
x' = \frac {M_{{10}}}{M_{{00}}}, y' = \frac {M_{{01}}}{M_{{00}}}
\end{equation}

The central moment is used to define translation invariants and makes their expressions more concisely.

The definitions expressed in Eq.~\eqref{eq:1} and Eq.~\eqref{eq:2} can be easily extended to higher dimensions or on different manifolds. For example, moments in 3D are

\begin{equation}
\begin{aligned}
\label{eq:4}
 M _{{\it lmn} }  &=  \int \!\!\! \int \!\!\! \int x ^{l  }y ^{m } z ^{n } \rho \left(x ,y ,z \right) d x  d y  d z \\
 \mu _{{\it lmn} }  &=  \int \!\!\! \int \!\!\! \int \left(x -x ' \right)^{l  }\left(y -y ' \right)^{m } \left(z -z ' \right)^{n } \\
 & \ \ \ \ \rho \left(x ,y ,z \right) d x  d y  d z \\
 x '   &=  \frac{M _{100}}{M _{000}}, y '   =  \frac{M _{010}}{M _{000}},z '   =  \frac{M _{001}}{M _{000}}
\end{aligned}
\end{equation}

Moments have a profound background in mathematics and physics, they are the coefficients of Fourier transform of images or shapes, containing all the information of them and can be used to reconstruct or recover images or shapes themselves. This property was formulated as the fundamental theorem \cite{Hu1962}.

\textbf{Definition 3: affine transform}

The affine transform or general linear transform in 2D is

\begin{equation}
\label{eq:5}
\mapleinline{inert}{2d}{u = x*a[1]+y*a[2]+a[3], v = x*b[1]+y*b[2]+b[3]}{$\displaystyle u=a_{{1}}x+a_{{2}}y+a_{{3}},\,v=b_{{1}}x+b_{{2}}y+b_{{3}}$}
\end{equation}

, or in matrix form

\begin{equation}
\label{eq:6}
\mapleinline{inert}{2d}{binomial(u, v) = T*binomial(x, y)+binomial(a[3], b[3])
 = Typesetting:-delayDotProduct(Matrix(
, binomial(x, y), true)+binomial(a[3], b[3])}{$\displaystyle {u\choose v}=T{x\choose y}+{a_{{3}}\choose b_{{3}}}
=  \left[ \begin {array}{cc} a_{{1}}&a_{{2}}\\ \noalign{\medskip}b_{{1}}&b_{{2}}\end {array} \right] {x\choose y} \\
\mbox{}+{a_{{3}}\choose b_{{3}}}$}
\end{equation}

Its Jacobi is

\begin{equation}
\label{eq:7}
\mapleinline{inert}{2d}{J = a[1]*b[2]-a[2]*b[1]}{$\displaystyle J=a_{{1}}b_{{2}}-a_{{2}}b_{{1}}$}
\end{equation}

when $a_3=b_3=0$, the transform is central affine transform.

\section{Moment invariants under affine group}
\label{sec:invariantsAffin}

The invariant has a long history \cite{Hilbert1994, ErlangenProgram}. For convenient to discussion in this paper, we adopt following

\textbf{Definition 4: invariant}

suppose there exists a function \mapleinline{inert}{2d}{(I)(a[1], a[2], () .. (), a[n])}{$\displaystyle I \left( a_{{1}},a_{{2}},{\ldots },a_{{n}} \right) $}
with parameters \mapleinline{inert}{2d}{a[j], j = 1, 2, () .. (), n}{$\displaystyle a_{{j}},\,j=1,\,2,\,{\ldots },\,n$}
, if the parameters \mapleinline{inert}{2d}{a[j]}{$\displaystyle a_{{j}}$}
are transformed to $a^{'}_{j}$ under some transform $T$ and

\begin{equation}
\label{eq:8}
I(a_{1}^{'},a_{2}^{'},...a_{n}^{'}) = w^{k}I(a_{1},a_{2},...a_{n})
\end{equation}

where $w$ is a function of transform parameters, $k$ is a real or integer as usual, then the function $I$ is called (relative) invariant (under transform $T$). When $k = 0$, it is called absolute invariant.

By moment invariants we mean some moment functions or rather homogeneous moment polynomials which are invariant under some transform groups. The degree of moment invariants is the degree of them being a polynomial.

The introduction of moment invariants to computer vision and image processing was proposed by \cite{Hu1962}, where seven moment invariants in 2D were given, which are normalized moment invariants and keep invariant under the transformations of translation, scale and rotation, or rather similarity transform group. They are widely used in areas of pattern recognition and shape analysis etc.

\begin{equation}
\begin{aligned}
\label{eq:9Hu}
I_1 &= \mu_{20} + \mu_{02} \\
I_2 &= (\mu_{20} - \mu_{02})^2 + 4 \mu_{11}^2 \\
I_3 &= (\mu_{30} - 3\mu_{12})^2 + (3\mu_{21} - \mu_{03})^2 \\
I_4 &= (\mu_{30} + \mu_{12})^2 + (\mu_{21} + \mu_{03})^2 \\
I_5 &= (\mu_{30} - 3\mu_{12})(\mu_{30} + \mu_{12})[(\mu_{30} + \mu_{12})^2 \\
& \ \ \ \ - 3(\mu_{21} + \mu_{03})^2] + (3\mu_{21} - \mu_{03})(\mu_{21} + \mu_{03}) \\
& \ \ \ \ * [3(\mu_{30} + \mu_{12})^2 - (\mu_{21} + \mu_{03})^2] \\
I_6 &= (\mu_{20} - \mu_{02})[(\mu_{30} + \mu_{12})^2 - (\mu_{21} + \mu_{03})^2] \\
& \ \ \ \ + 4\mu_{11}(\mu_{30} + \mu_{12})(\mu_{21} + \mu_{03}) \\
I_7 &= (3\mu_{21} - \mu_{03})(\mu_{30} + \mu_{12})[(\mu_{30} + \mu_{12})^2 \\
& \ \ \ \ - 3(\mu_{21} + \mu_{03})^2] -(\mu_{30} - 3 \mu_{12})(\mu_{21} + \mu_{03}) \\
& \ \ \ \ *[3(\mu_{30} + \mu_{12})^2 - (\mu_{21} + \mu_{03})^2]
\end{aligned}
\end{equation}

Moment invariants under affine transform was proposed by \cite{Flusser1993}, where four invariants were explicitly given. Later, a graph method to describe the construction visually was proposed by \cite{Suk_Flusser2011} and \cite{Flusser2017}. Three low order affine moment invariants are

\begin{equation}
\begin{aligned}
\label{eq:10}
I_1 &= (\mu_{20} \mu_{02} - \mu_{11}^2) / \mu_{00}^4 \\
I_2 &= (\mu_{30}^2 \mu_{03}^2 - 6 \mu_{30}\mu_{21}\mu_{12}\mu_{03} + 4 \mu_{30}\mu_{12}^3  \\
& \ \ \ \ + 4 \mu_{21}^3 \mu_{03} - 3 \mu_{21}^2 \mu_{12}^2) / \mu_{00}^{10} \\
I_3 &= (\mu_{20}(\mu_{21}\mu_{03} - \mu_{12}^2) - \mu_{11}(\mu_{30}\mu_{03} - \mu_{21}\mu_{12}) \\ 
& \ \ \ \ + \mu_{02}(\mu_{30}\mu_{12} - \mu_{21}^2)) / \mu_{00}^7 \\
\end{aligned}
\end{equation}

And moment invariants in 3D were first developed by \cite{Sadjadi1980}, where three invariants were given, which are the three sequential principal minors of a matrix composed of second order moments.

\begin{equation}
\begin{aligned}
\label{eq:11}
I_1 &= \mu_{200} + \mu_{020} + \mu_{002} \\
I_2 &= \mu_{200}\mu_{020} + \mu_{200}\mu_{002} + \mu_{020}\mu_{002} - \mu_{110}^2 \\
& \ \ \ \ - \mu_{101}^2 - \mu_{011}^2 \\
I_3 &= \mu_{200}\mu_{020}\mu_{002} + 2\mu_{110}\mu_{101}\mu_{011} - \mu_{002}\mu_{110}^2 \\
& \ \ \ \ - \mu_{020}\mu_{101}^2  - \mu_{002}\mu_{011}^2 \\
\end{aligned}
\end{equation}

Moment invariants in N dimensional space was expressed by \cite{Mamistvalov98}.

One of basic methods for deriving moment invariants is from the theory of algebraic invariant \cite{Hu1962}, \cite{Sadjadi1980}, \cite{Flusser1993}. Another is from group theory or sphere harmonics in terms of complex moments \cite{LoD89}.

The multiple integrals of geometric invariant cores proposed by \cite{Li2006, Xu2006,XuL2006, Xu2008} provided a general and intuitive way to build moment invariants with different order and degree in any dimensions or on manifolds.

According to this method, images and shapes can be seen to be attached a structure of geometric entities or geometric primitives, like distance, angle, area, and volume etc, which form a geometric core function that is a multiply of the entities and may keep invariant under some geometric transforms. For example, the distance is invariant under isometric transform, the area and the volume are relative invariant under affine transform. Moment invariants can then be defined by multiple integrals with the degree depending on the number of vector points involved and the order depending on the frequency of points appearing in the definition. In this way, the definition of moment invariants is conveniently extended to higher dimension of spaces and to manifolds, like curves and surfaces \cite{Li2006, Xu2006, XuL2006, Xu2008}.

Concretely, geometric primitives for definition of central moment invariants in 2D are

\begin{equation}
\begin{aligned}
\label{eq:12_2-D primitive}
D(O,i) &=  (x^{2}_{i} + y^{2}_{i})^{\frac{1}{2}} \\
R(O,i,j) &= (x_{i},y_{i}) \cdot (x_{j},y_{j}) \\
A(O,i,j) &=  \frac{1}{2} (\left\|(x_{i},y_{i})\times(x_{j},y_{j}) \right\|)^{2}
\end{aligned}
\end{equation}

where $D, R$ and $A$ represent the distance between two points, the angle between two vectors, and the triangular area formed by three vertices, respectively.

For example, taking $D^2$ to be a core function, Hu's first invariant Eq.~\eqref{eq:9Hu} can be simply calculated as

\begin{equation}
\label{eq:13}
\mapleinline{inert}{2d}{I[1]*(int(int((x[1]^2+y[1]^2)*l(x[1], y[1]), x[1]), y[1])) = mu[20]+mu[2]}{$\displaystyle I_{{1}} = \int \!\!\!\int \! \left( {x_{{1}}}^{2}+{y_{{1}}}^{2} \right) \rho \left( x_{{1}},y_{{1}} \right) {dx_{{1}}}\,{dy_{{1}}}\\
\mbox{}=\mu_{{20}}+\mu_{{02}}$}
\end{equation}

where $l(x, y)$ is the density function of image.

All moment invariants available currently in the literature can be expressed as multiple integrals of geometric cores, and their defining geometric primitives are easily recovered \cite{Xu2008}.

Careful analysis of geometric primitives shows that they can be simplified to two basic functions: vector inner product and vector cross product or determinant, which were called generating functions. Such a simplification provides a new view angle to the intrinsic structure of invariant \cite{Li2017}.

\begin{equation}
\label{eq:14_generatingFunc}
f(i,j) = \sum^m_{k=1} x_{ik}x_{jk}
\end{equation}

\begin{equation}
\label{eq:14_generatingFunc2}
g(i_1,i_2,...i_m) = det \left|
\begin{array}{cccc}
x_{i11} & x_{i12} & \cdots & x_{i1m} \\
x_{i21} & x_{i22} & \cdots & x_{i2m} \\
\cdots  & \cdots  & \cdots & \cdots \\
x_{im1} & x_{im2} & \cdots & x_{imm} \\
\end{array} \right|
\end{equation}

where $f$ is two vector inner product, and $g$ is the determinant of order m by m matrix composed of vectors.

For example, the well-known Hu's seven invariants expressed in generating functions are as following:

\begin{equation}
\begin{aligned}
\label{eq:15_Hu-generate}
I_1 &\Leftrightarrow f(1,1) \\ 
I_2 &\Leftrightarrow (f(1,2))^2 - 2(g(1,2))^2 \\
I_3 &\Leftrightarrow (f(1,2))^3 - 3(g(1,2))^2f(1,2) \\
I_4 &\Leftrightarrow f(1,2)f(1,1)f(2,2) \\
I_5 &\Leftrightarrow f(2,2)f(3,3)f(4,4)[f(2,1)f(3,1)f(4,1) \\
& \ \ \ \ -f(2,1)g(3, 1)g(4,1)-g(2,1)g(3,1)f(4,1) \\
& \ \ \ \ -g(2,1)f(3,1)g(4,1)] \\
I_6 &\Leftrightarrow f(2,2)f(3,3)[f(1,2)f(1,3) - g(1,2)g(1,3)] \\
I_7 &\Leftrightarrow f(2,2)f(3,3)f(4,4)[g(2,1)f(3,1)f(4,1) \\
& \ \ \ \ -g(2,1)g(3,1)g(4,1)+f(2,1)g(3,1)f(4,1) \\
& \ \ \ \ +f(2,1)f(3,1)g(4,1)]
\end{aligned}
\end{equation}

where $f(i,j) = x_{i}x_{j} + y_{i}y_{j}, g(i,j) = x_{i}y_{j} - y_{i}x_{j}$.
A simple observation is that Hu's seven invariants are not simplest, they are polynomials and can be further split into more simple ones, the monomials consisting of multiplication of two generating functions f and g. Those monomials were more fundamental and called primitive invariants (PIs) or ShapeDNA, because they convey the character of images or shapes and the deriving procedure is just like DNA encoding the protein in biology. The rule converting from multiplication of generating functions to invariants is simple and clear, that is the invariant degree depends on the number of vector points involved in generating functions, the power of coordinates is translated into the order of moments, which depends on the frequency of points appearing in the functions.

For instance, $I_{2}$ in Eq.~\eqref{eq:15_Hu-generate} is composed of two primitives $f(1, 2)^2 $
and $g(1, 2)^2 $. For the second one, there are two points involved,whose frequencies are also two. Thus two sets of integrals are needed. We first expand the generating function as

\begin{equation}
\label{eq:16}
\mapleinline{inert}{2d}{x[1]^2*y[2]^2-2*x[1]*x[2]*y[1]*y[2]+x[2]^2*y[1]^2}{$\displaystyle {x_{{1}}}^{2}{y_{{2}}}^{2}-2\,x_{{1}}x_{{2}}y_{{1}}y_{{2}}+{x_{{2}}}^{2}{y_{{1}}}^{2}$}
\end{equation}

and then two sets of double integrals are calculated


\begin{equation}
\label{eq:17}
\begin{aligned}
\int \!\!\!\int \! & ( \int \!\!\! \int \! ( {x_{{1}}}^{2}{y_{{2}}}^{2} - 2 x_{{1}}x_{{2}}y_{{1}}y_{{2}} + {x_{{2}}}^{2}{y_{{1}}}^{2} ) \\
& \rho ( x_{{1}},y_{{1}} ) {dx_{{1}}}\,{dy_{{1}}} ) \rho ( x_{{2}},y_{{2}} ) {dx_{{2}}}{dy_{{2}}} \\
& = 2 ( \mu_{{20}}\mu_{{2}} - {\mu_{{11}}}^{2})
\end{aligned}
\end{equation}

Noticed that the constant in the primitives or invariants is often omitted for simplicity.

Therefore, Hu's seven invariants are in fact composed of sixteen PIs. A simplest set of lower order and lower degree PIs is proposed in \cite{Li2017}.

It is easy to check that invariants composed of only $f$ functions are isometric invariants, including rotation and uniform scaling by a proper normalization,while those of only $g$ functions are affine invariants under general linear transform.

For the general linear transform or affine transform in Eq.~\eqref{eq:6}, there is a relation of its Jacobi in Eq.~\eqref{eq:7} with two zero order moments. Let $M_{00}$ and $N_{00}$ denote the moments before and after transform.

\begin{equation}
\label{eq:18}
\begin{aligned}
M_{{00}} &= \int \!\!\!\int \!\rho \left( u,v \right) {du}\,{dv} = \int \!\!\!\int \!\rho \left( x,y \right)J{dx}\,{dy} \\
& = J\int \!\!\!\int \!\rho \left( x,y \right) {dx}\,{dy} = {J}\ N_{00} \\
{J} &= {\frac {M_{{00}}}{N_{{00}}}}
\end{aligned}
\end{equation}

Thus the powers in the denominators of Eq.~\eqref{eq:10} depends on two factors, one is the number of $g$ 
functions involved in the definition, and another is the number of vector points, which relates to the multiplicity of integrals. As result, when Eq.~\eqref{eq:17} defines an affine invariant, it has power four in the denominator as shown in Eq.~\eqref{eq:10}. 

We have the following algorithm
\\

\noindent\rule{8cm}{0.4pt}
\textbf{Algorithm 1: derivation of PIs } \\
\noindent\rule{8cm}{0.4pt}

Given generating functions $f$ and $g$, vector points $p_{j}$, \\

\textbf{1.} Choose $m$, the vector point number involved, which determines the degree of PI; \\

\textbf{2.} Choose $n_j, j = 1, 2, ... , m$, the frequency for each $p_{j}$, which will appear in function $f$
or $g$; $n_j$ will determine the order of the moment defined by $p_{j}$; \\

\textbf{3.} Construct $f$ and $g$, and determine $l$ and $k$, the number of $f$ and $g$ involved; if an affine invariant is desired, only function $g$ is to be used, and the number of $g$ functions is at least even; \\

\textbf{4.} Construct a multiplication of $f$ and $g$ to form a definition of PI; \\

\textbf{5.} Expand the multiplication as a polynomial of coordinates variants; \\

\textbf{6.} Translate the power of variants to the order of moments; \\
if the construction is not zero, continue; \\
else go to 1 and try again; \\

\textbf{7.} Simplify and omit the constant factor, if any; for an affine PI, a denominator of zero order moment with power $m+k$ is to be added.

\noindent\rule{8cm}{0.4pt} \\

Notice that some constructions of PIs may be zero finally, because one cannot differentiate if it is effective before the "encoding" or translating really happens. Since the number of combinations is not large for limited degree or order, one can always get what required quickly, see the following examples.

The algorithm is general. For N dimensional space, $f$ is $N$ dimensional vector inner product and $g$ is the determinant of order $N * N$  matrix. For curves and surfaces, adaptively correspondent curve or surface integrals are needed for the definition.

\section{Moments relationships under geometric transforms}
\label{sec:relationships}

The invariants discussed till now are all group invariants, concrete transform parameters are canceled out. And the invariants for each group are the same, which are all defined in central moments. Of course, all the central moments are invariant for translation group. Besides, the invariants for groups of rotation, similarity,and affine are defined. Geometrically, the affine group contains similarity, and the later contains rotation.

The relationship between moments under transform Eq.~\eqref{eq:6} is trivial. A simple fact is that when central moment is chosen, the first order moments are all zero. Let $M_{jk}$ and $N_{jk}$ denote the moments before and after transform, for the second order moments, we have

\begin{equation}
\begin{aligned}
\label{eq:19}
M_{20} &= a_{1}^2N_{20} + 2a_{1}a_{2}N_{11} + a_{2}^2N_{02} \\
M_{11} &= a_{1}b_{1}N_{20} + (a_{1}b_{2} + a_{2}b_{1})N_{11} + a_{2}b_{2}N_{02} \\
M_{02} &= b_{1}^2N_{20} + 2b_{1}b_{2}N_{11} + b_{2}^2N_{02} \\
\end{aligned}
\end{equation}

or in matrix form

\begin{equation}
\label{eq:20}
\left|
\begin{array}{c}
M_{20} \\
\\
M_{11} \\
\\
M_{02} \\
\end{array} \right|
= 
\left|
\begin{array}{ccc}
a_{1}^2 & 2a_{1}a_{2} & a_{2}^2 \\
\\
a_{1}b_{1} &(a_{1}b_{2} + a_{2}b_{1}) & a_{2}b_{2} \\
\\
b_{1}^2 & 2b_{1}b_{2} & b_{2}^2  \\
\end{array} \right|
\left|
\begin{array}{c}
N_{20} \\
\\
N_{11} \\
\\
N_{02} \\
\end{array} \right|
\end{equation}

This relation keeps true for any order moment in $N$ dimension. We have

\textbf{Proposition 1:}There exists a linear relation among the same order moments under affine transform.

\section{From integral invariants to differential invariants }
\label{sec:integralToDifferential}

The derivation of differential invariants is usually difficult or complicated \cite{Guggenheimer1963}. The general theory and technique of Lie group and Lie algebra were developed by \cite{Lie1884}. Moving frame was established by \cite{olver1995,Olver1999,Olver2007, Olver2009, Olver2010}. And recently, \cite{Wang2013} explicitly gave 11 affine differential invariants in 2D by using theory of binary forms. From moment invariant, the differential invariant can find a more intuitive way to derive.

For a 2D function $H(u(x,y), v(x,y))$, the relations of the second order derivatives under transform Eq.~\eqref{eq:6} are

\begin{equation}
\begin{aligned}
\label{eq:21}
\frac{\partial^2 H}{\partial x^2} &= a_{1}^2 \frac{\partial^2 H}{\partial u^2} + 2a_{1}b_{1} 
\frac{\partial^2 H}{\partial u \partial v} + b_{1}^2 \frac{\partial^2 H}{\partial v^2} \\
\frac{\partial^2 H}{\partial x \partial y} &= a_{1}a_{2} \frac{\partial^2 H}{\partial u^2} + (a_{1}b_{2} + b_{1}a_{2}) \frac{\partial^2 H}{\partial u \partial v} + b_{1}b_{2} \frac{\partial^2 H}{\partial v^2} \\
\frac{\partial^2 H}{\partial y^2} &= a_{2}^2 \frac{\partial^2 H}{\partial u^2} + 2a_{2}b_{2} 
\frac{\partial^2 H}{\partial u \partial v} + b_{2}^2 \frac{\partial^2 H}{\partial v^2} \\
\end{aligned}
\end{equation}

or in matrix form

\begin{equation}
\label{eq:22}
\left|
\begin{array}{c}
\frac{\partial^2 H}{\partial x^2} \\
\\
\frac{\partial^2 H}{\partial x \partial y} \\
\\
\frac{\partial^2 H}{\partial y^2} \\
\end{array} \right|
= 
\left|
\begin{array}{ccc}
a_{1}^2 & 2a_{1}b_{1} & b_{1}^2 \\
\\
a_{1}a_{2} &(a_{1}b_{2} + a_{2}b_{1}) & b_{1}b_{2} \\
\\
a_{2}^2 & 2a_{2}b_{2} & b_{2}^2  \\
\end{array} \right|
\left|
\begin{array}{c}
\frac{\partial^2 H}{\partial u^2} \\
\\
\frac{\partial^2 H}{\partial u \partial v } \\
\\
\frac{\partial^2 H}{\partial v^2} \\
\end{array} \right|
\end{equation}

That is the second order derivatives fit for an affine transform. The result is true for any order in $N$ dimension. We have

\textbf{Proposition 2:}There exists a linear relation among the same order derivatives under affine transform.

As discussed above, the moments have the same relation and a direct reasoning is

\textbf{Proposition 3:}There exists an isomorphism between differential invariants and moment invariant in $N-D$ or manifolds.

\textbf{Corollary 1:} The two generating functions $f$ and $g$ for moment invariants are also those for differential invariants.

\textbf{Corollary 2:}The differential invariants under rotation transform group can be obtained by moment invariants under the same transform group by translating the moments into the same order derivatives.

\textbf{Corollary 3:}The differential invariants under affine transform group can be obtained by moment invariants under the same transform group by translating the moments into the same order derivatives, with a revision in denominator to $J_k$, where $J$ is the Jacobi of the transform, $k$ is the number of $g$ functions involved.

For example, the first invariant in Eq.~\eqref{eq:9Hu} will be a Laplace operator when translated to derivatives, and it is rotation invariant or isotropical; the first invariant in Eq.~\eqref{eq:10} will be affine differential invariant if translated to derivatives in numerator, and $J_2$ in place of the denominator.

\textbf{Corollary 4:}The differential invariants containing the first order derivatives are also meaningful, for they are no longer zero.

For example, as \cite{olver1995} indicated, the following is an affine differential invariant with order two (the power of Jacobi ),

\begin{equation}
\label{eq:23}
\frac{1}{J^2}( \frac{\partial^2 H}{\partial x^2}( \frac{\partial H}{\partial y})^2 
- 2 \frac{\partial H}{\partial x} \frac{\partial H}{\partial y} \frac{\partial^2 H}{\partial x \partial y} 
+ \frac{\partial^2 H}{\partial y^2}( \frac{\partial H}{\partial x})^2)
\end{equation}

and the analysis to it shows that it is a three degree polynomial in derivatives, and the power of the denominator is two. According to the generating rule above, its generating functions are a multiplication of two $g$ functions with three points involved. We can try

\begin{equation}
\label{eq:24}
g(2,1)g(2,3)
\end{equation}

and directly expanding goes

\begin{equation}
\label{eq:25}
\mapleinline{inert}{2d}{-x[1]*x[2]*y[2]*y[3]+x[1]*x[3]*y[2]^2+x[2]^2*y[1]*y[3]-x[2]*x[3]*y[1]*y[2]}{$\displaystyle -x_{{1}}x_{{2}}y_{{2}}y_{{3}}+x_{{1}}x_{{3}}{y_{{2}}}^{2}+{x_{{2}}}^{2}y_{{1}}y_{{3}}\\
\mbox{}-x_{{2}}x_{{3}}y_{{1}}y_{{2}}$}
\end{equation}

translating into moment gets

\begin{equation}
\label{eq:26}
\mu_{20}\mu_{01}^2 - 2\mu_{11}\mu_{10}\mu_{01} + \mu_{02}\mu_{10}^2 
\end{equation}

Finally, the differential invariant in Eq.~\eqref{eq:23} is fixed by replacing moments with the same order derivatives.

\section{Extension to projective differential invariants }
\label{sec:projective}

Geometrically, projective transform group contains affine transformations. Directly finding a projective differential invariant is not trivial \cite{olver1995,Olver2010}. While affine differential invariants can be well established, projective differential invariants could be screened out from affine ones.

Testing shows the differential invariant in Eq.~\eqref{eq:23} is also a projective one, with its generating function in Eq.~\eqref{eq:24}. Similarly in 3D, if a generating function is designed as

\begin{equation}
\label{eq:27}
\frac{1}{2}g(1,2,3)g(2,3,4)
\end{equation}

it means a construction of degree 4, including a multiplication of two first order derivatives and two second order derivatives in each item, and power two in the denominator. In fact, it defines a projective differential invariant

\begin{equation}
\begin{aligned}
\label{eq:28}
& \frac{1}{J_3^2}(I_x^2(I_{yy}I_{zz} - I_{yz}^2) + I_y^2(I_{xx}I_{zz} - I_{xy}^2) + \\
& I_z^2(I_{xx}I_{yy} - I_{xy}^2) + 2I_{x}I_{y}(I_{xz}I_{yz} - I_{xy}I_{zz}) + \\
& 2I_{y}I_{z}(I_{xy}I_{xz} - I_{yz}I_{xx}) + 2I_{x}I_{z}(I_{xy}I_{yz} - I_{xz}I_{yy}))
\end{aligned}
\end{equation}

where $J_3$ is the transform Jacobi in 3D. 

Eq.~\eqref{eq:28} was first given in \cite{Rathi2006} as an affine invariant.

Compare Eq.~\eqref{eq:27} and Eq.~\eqref{eq:24}, we have 

\textbf{Conjecture 1:}The following generating function would define a projective differential invariant in $m-D$, except a sign or constant:

\begin{equation}
\label{eq:29}
g(1,2,...,m-1)g(2,3,...,m)
\end{equation}

\section{Conclusions}
This paper reports a discovery that under the affine transform, differential invariants have similar structures with moment invariants up to a scalar function of transform parameters. If moment invariants are known, relative differential invariants can be obtained by the substitution of moments by derivatives with the same order. 

Whereas moment invariants can be calculated by multiple integrals, this method provides a simple way to derive differential invariants without the need to resolve any equation system.

Since the definition of moments on different manifolds or in different dimension of spaces is well established, differential invariants on or in them are also well defined. Considering that moments have a strong background in mathematics and physics, this technique offers a new view angle to the inner structure of invariants.

Projective differential invariants can also be found in this way with a screening process.


%


\section*{Acknowledgment}

This work was partly funded by National Natural Science Foundation of China (Grant No. 60873164, 61227082 and  61379082).

\ifCLASSOPTIONcaptionsoff
  \newpage
\fi


\bibliographystyle{IEEEtran}
\bibliography{Differential}
\end{document}